\documentclass[11pt]{article}

\usepackage{arxiv}

\usepackage[utf8]{inputenc}
\usepackage[T1]{fontenc}
\usepackage{microtype}

\usepackage{amsmath,amssymb,amsfonts}
\usepackage{newpxtext}   % Palatino-family body font
\usepackage{newpxmath}   % matching math font
\usepackage{bm}

\usepackage{booktabs}
\usepackage{multirow}
\usepackage{array}
\usepackage{float}   % enables [H] to pin tables/figures exactly in the text

\usepackage{graphicx}
\usepackage{tikz}
\usetikzlibrary{arrows.meta, positioning, shapes.geometric, calc, fit, backgrounds}
\usepackage{caption}
\DeclareCaptionLabelSeparator{vbar}{ $|$ }
\usepackage{subcaption}
\graphicspath{{output/}{figures/}}

\usepackage{enumitem}
\usepackage{xcolor}
\usepackage{url}

\usepackage[numbers,sort&compress]{natbib}
\usepackage{hyperref}
\usepackage[capitalise,noabbrev]{cleveref}

\newcommand{\nds}{NDS\xspace}                 % Nutrition Data Service
\newcommand{\ndsfull}{Nutrition Data Service\xspace}

\newcommand{\nutribench}{NutriBench\xspace}

\newcommand{\sssom}{SSSOM\xspace}
\newcommand{\skos}{SKOS\xspace}
\newcommand{\exactmatch}{\texttt{skos:exactMatch}\xspace}
\newcommand{\broadmatch}{\texttt{skos:broadMatch}\xspace}

\newcommand{\fone}{F$_1$\xspace}

\usepackage{xspace}

\title{Nutrition Data Infrastructure for the AI Era:\\
Operationalizing FAIR for Agent-Mediated Research}

\author{%
  \begin{tabular}{c}
    Lin Liao \\[3pt]
    {\small \texttt{lin@8up.ai}}
  \end{tabular}%
  \hspace{2em}% arxiv.sty's \And separator is clobbered by amsmath's math-mode \And
  \begin{tabular}{c}
    Peng Li \\[3pt]
    {\small \texttt{peng@8up.ai}}
  \end{tabular}%
}

\date{\today}
\runningtitle{Nutrition Data Infrastructure for the AI Era}

\begin{document}
\maketitle

\begin{abstract}
AI agents can accelerate nutrition research, but their analyses inherit the
identity, semantic, and release ambiguities of the underlying data. We present
\ndsfull (\nds), source-preserving infrastructure that operationalizes FAIR for
automated use: description resolution makes release-specific records findable;
typed crosswalks connect independently released resources;
machine-readable interfaces expose versioned sources and crosswalks, supporting
replayable and auditable analyses. On food-description benchmarks, \nds
outperforms the best published
language-model result on NutriBench. External and blinded crosswalk evaluations
show that its typed contract favors defensible links and rejects unsupported
mappings. In a person-level glycemic-index analysis, pinned \nds inputs produce
identical outputs across models and repeated runs, while open-web reconstruction
remains unstable. Together, these results show that agent-mediated nutrition
research requires a new infrastructure that makes data identity, search, and
crosswalk policy explicit.
\end{abstract}

\section{Introduction}
\label{sec:intro}

AI agents can retrieve literature, operate research tools, write and execute
analysis code, and combine evidence across sources. We use
\emph{agent-mediated research} for a human-directed workflow in which an agent
performs bounded operations---finding evidence, reconciling datasets, running
analyses, and recording artifacts---while the researcher sets the question,
constraints, and scientific interpretation~\citep{xin2025agentic}. Its promise
is not merely faster answers, but a shorter path from hypothesis to auditable
sensitivity analysis.

Nutrition makes both the opportunity and the constraint unusually clear. A
single study may span food-composition tables, dietary surveys,
branded-product catalogs, prices, biomarkers, and health outcomes. An agent must
not only retrieve a plausible nutrient value; it must select the intended food
record, preserve its source and release, interpret analytical bases, and make
cross-source joins that remain auditable. Retrieval grounding reduces reliance
on parametric memory, but it cannot repair ambiguous identifiers or missing
context in the evidence itself~\citep{lewis2020rag,ji2023hallucination}.

The present food-data landscape makes those failures likely. An assessment of
101 food-composition databases covering 110 countries found that only 32\%
offered an API and only 17 satisfied all 13 evaluated criteria of the
FAIR principles (findable, accessible, interoperable, and
reusable)~\citep{brinkley2025state}. The review also reports that
nutrient-intake estimates for an identical diet can vary by
20--45\% with the database selected~\citep{brinkley2025state}.
Interoperability is especially difficult because no database can provide it
alone~\citep{li2023quality,jenningsdobbs2023interop}.
Food sources rarely share stable identifiers, while names vary with geography,
species, preparation, edible portion, processing, brand, and database purpose.
A string join can silently conflate nutritionally distinct foods; flattening
sources into one table can erase the very release history and analytical
semantics needed to detect the mistake. Bridging independently released
sources instead requires an explicit mapping assertion: which records are
related, by what relation, with what evidence and confidence, under which
endpoint releases, and with an explicit unsupported outcome when no defensible
link exists. Description-driven food matching is thus not an application added
to the infrastructure; it is the mechanism that makes otherwise disconnected
sources interoperable.

We therefore use FAIR as the requirements vocabulary. AI does not redefine
FAIR; it raises the standard for a sufficient implementation. Context that a
careful researcher might recover from documentation must be machine-actionable,
and ambiguous joins must fail visibly rather than be improvised at query time.

We present \ndsfull (\nds), source-preserving infrastructure
built toward these requirements. \nds retains independently addressable source
records and nutrient observations; resolves natural-language descriptions with
hybrid retrieval, contextual ranking, and abstention; constructs auditable
crosswalks without collapsing their endpoints; and exposes structured data
through REST, bulk export, and Model Context Protocol (MCP) tools.

We make three contributions. \textbf{(1)~Infrastructure:} a deployed data model
and access layer that retain source identity, release, nutrient semantics, and
stored lineage. \textbf{(2)~Resolution and crosswalks:} a shared
description-driven pipeline for natural-language lookup and cross-source
mapping, demonstrated in a deployed crosswalk from the Food and Nutrient
Database for Dietary Studies (FNDDS) to glycemic-index (GI) records.
\textbf{(3)~Evaluation:} tests
of record resolution, end-to-end nutrient estimation, open-set alignment,
crosswalk quality, and repeatability under agent-mediated use, with source reconciliation as
supporting validation.

\Cref{sec:related} discusses related work,
\Cref{sec:fair-ai} derives the AI-era FAIR criteria; \cref{sec:method}
describes the system; and \cref{sec:experiments,sec:limitations} report the
evidence and its limits.

\section{Related Work}
\label{sec:related}

\paragraph{Food-composition data and harmonization.}
Food-composition databases are foundational to nutrition science, yet their
fragmentation and uneven stewardship are well documented. Global assessments
identify infrequent updates, extensive reuse of secondary data, and geographic
disparities in FAIR adoption~\citep{brinkley2025state}; published studies also
often omit the database version needed for reproducibility~\citep{li2023quality}.
Crosswalks are sparse across composition, price, environmental, and geospatial
sources~\citep{jenningsdobbs2023interop}. EuroFIR and emerging
minimum-information standards address harmonization at compilation time
\citep{pakkala2010harmonised,blumberg2025call}. We instead connect
independently released composition, survey, and branded-food sources that did
not adopt a common standard~\citep{fukagawa2022fdc}.

\paragraph{Food semantics and mapping standards.}
FoodOn transforms much of LanguaL's vocabulary into an OWL ontology with
explicit relations among food products, ingredients, qualities, and
processes~\citep{dooley2018foodon}. The Simple Standard for Sharing Ontological
Mappings (SSSOM) provides a complementary schema for
exchanging mappings with typed relations, justification, provenance, and
confidence metadata~\citep{sssom2022}. These standards make foods and mappings
more machine-readable, but do not determine which record-level mappings to
assert, when to abstain, or how to maintain mappings as source releases change.

\paragraph{Food record matching.}
Prior systems map food descriptions using fuzzy distances, embeddings, and
large language models (LLMs). \citet{moralesgarzon2020embedding} show that one
description may admit
defensible candidates at several levels of specificity. NutriBench evaluates
carbohydrate estimation from meal descriptions~\citep{nutribench2025}.
\citet{lemay2026mapping} make abstention a first-class outcome and show that
similarity thresholds do not cleanly separate matches from non-matches.
NutriMatch~\citep{jankelow2026nutrimatch}
combines LLM normalization, embedding retrieval, and LLM validation
to expand nutrient coverage across national databases.
These systems motivate our focus on matching
as infrastructure: not only selecting a record, but publishing the relation,
evidence, release scope, and abstention needed to audit or replay that choice.

\paragraph{LLMs as the nutrition access layer.}
Benchmarks such as NGQA now test personalized reasoning over National Health
and Nutrition Examination Survey (NHANES) profiles and FNDDS
foods~\citep{ngqa2025}. LLMs are also being evaluated as direct interfaces for
guideline-adherent nutrition
information~\citep{parameswaran2025nutrition}, raising the stakes for grounding
answers in versioned, source-preserving data. Yet model answers are difficult
to cite or replay and remain prone to
hallucination~\citep{ji2023hallucination}.
Retrieval grounding~\citep{lewis2020rag} helps only when the interface preserves
the source, release, and mapping context of the evidence it returns.

\section{FAIR Principles in the AI Era}
\label{sec:fair-ai}

FAIR addresses both human researchers and computational agents. It describes
properties of digital research objects---including data, metadata, algorithms,
and workflows---without prescribing a storage or search
architecture~\citep{wilkinson2016fair}. Applying it therefore requires
domain-specific
interpretation~\citep{jacobsen2020fair}.

For agent-mediated nutrition research, the relevant objects range from a
dataset release to a record, value, or mapping. \Cref{tab:fair-mapping} states
the criteria used in this paper and the corresponding NDS mechanisms.

\begin{table}[H]
  \centering
  \small
  \caption{FAIR criteria and NDS mechanisms for agent-mediated nutrition
  research.}
  \label{tab:fair-mapping}
  \begin{tabular}{@{}
    >{\raggedright\arraybackslash}p{0.14\linewidth}
    >{\raggedright\arraybackslash}p{0.40\linewidth}
    >{\raggedright\arraybackslash}p{0.38\linewidth}@{}}
    \toprule
    FAIR requirement & Agent-mediated criterion & NDS mechanism \\
    \midrule
    Findable &
    An agent can discover the intended release, record, or mapping by identifier
    or description, without confusing discovery with identity. &
    Source-derived identifiers, release metadata, indexed fields, hybrid
    description resolution, and explicit unsupported results. \\
    \addlinespace[2pt]
    Accessible &
    An agent can retrieve data and interpretive metadata by identifier through
    machine-readable, authentication-aware interfaces. &
    REST, bulk export, and MCP interfaces preserve source and release identity
    in identifier-based retrieval. \\
    \addlinespace[2pt]
    Interoperable &
    An agent can combine independent objects through shared representations and
    qualified references without inventing unsupported relationships. &
    Shared food and nutrient models, preserved source semantics, typed mappings,
    abstention, and standards-oriented alignment exports.
    \\
    \addlinespace[2pt]
    Reusable &
    An agent can judge whether and how an object may be reused from its
    semantics, constraints, provenance, and version. &
    Preserved units, bases, and absent-versus-zero observations; stored source
    and transformation lineage; and pinned mapping resolution. \\
    \bottomrule
  \end{tabular}
\end{table}

\section{System}
\label{sec:method}

\subsection{Architecture}
\label{sec:architecture}

\begin{figure}[H]
  \centering
  \includegraphics[width=\linewidth]{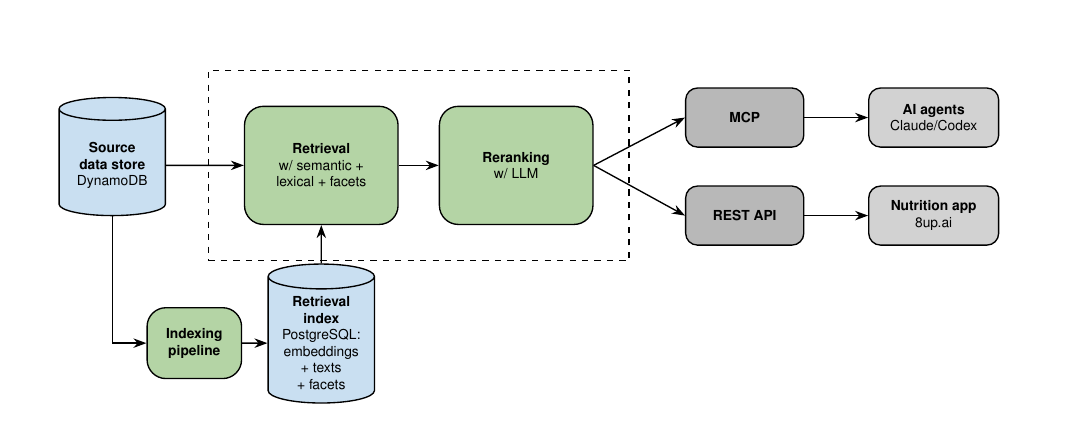}
  \caption{Nutrition Data Service (NDS) separates authoritative source storage, a rebuildable retrieval
  index, and access for AI agents and applications.}
  \label{fig:architecture}
\end{figure}

\cref{fig:architecture} shows a high-level architecture of NDS. Offline indexing builds a
search index from source descriptions. Online requests use that index to
retrieve and rerank candidates, then expose selected source records through
MCP or REST. MCP is an open protocol through which agents discover and invoke
named external tools~\citep{mcp2024}. Authoritative values remain in the source
store; the retrieval index is a rebuildable aid to discovery, not scientific
evidence.

The figure focuses on food matching; ingestion, caching, and bulk export are
omitted for clarity.

\subsection{Storage}
\label{sec:storage}
NDS imports heterogeneous sources into DynamoDB while preserving records that
appear to describe the same food. Each food receives the deterministic
key
$\mathit{food\_uid}=\textsc{uuid5}(\textit{source\_id},
\textit{source\_record\_id})$, so re-importing a release reproduces the same
identity. The record also carries its source system, dataset, and release as
explicit fields; for example, an FNDDS record names both its survey cycle and
the FoodData Central release from which it was imported.

Physical tables separate non-branded foods, branded products, surveys,
indicators and references, and crosswalk records. This avoids forcing
very different access patterns into one schema while retaining stable links
among records.

Food matching uses PostgreSQL with pgvector as a separate retrieval index.
Each description has an embedding for semantic similarity, a normalized name
for lexical search, and structured facets for nutritionally meaningful traits:
base food, cooked state, cooking method, form, preservation, coating, dish
type, and preparation additives.

\subsection{Description-driven food matching}
\label{sec:matching}

When an identifier is known, NDS retrieves the record directly. Otherwise it
uses the name or description common to most sources. The English-language path
has three stages: parsing, high-recall retrieval, and precision-oriented
reranking.

\paragraph{Query parsing.}
An LLM constrained by a structured schema decomposes the description into the
same normalized base food and closed-vocabulary facets used in the index. On a
parse failure, the system falls back to the normalized description.

\paragraph{Hybrid retrieval.}
Two channels nominate candidates independently. The semantic channel retrieves
nearby base-food embeddings and scores cosine similarity adjusted for specified
facet agreement; the deployed weighting is
$0.70\cdot\mathit{cosine} + 0.30\cdot\mathit{facet}$, with the facet term
normalized over facets present in the query. Unknown candidate facets receive
half credit, while unspecified query facets are ignored. The lexical channel
full-text searches normalized names. Reciprocal rank fusion combines the lists:
a candidate receives $1/(k+r)$ for rank $r$ in each returning channel
(deployed $k{=}60$), favoring records found by both. The reranker receives the
top fused candidates.

\paragraph{LLM-based reranking.}
One listwise LLM call compares the candidates using the original description,
candidate names, and facets, with emphasis on preparation, physical form,
defining ingredients, and brand. It returns a verdict, score, and ordering.

\paragraph{Configuration.}
The evaluated deployment parses and reranks with \texttt{gemini-3.6-flash}
and uses 1{,}536-dimensional \texttt{gemini-embedding-2-preview} vectors in an
hierarchical navigable small-world (HNSW) cosine index. The semantic and lexical
channels retrieve up to 250 and 25
records, respectively; the reranker receives the top 25 fused candidates and
returns at most 5.

If no candidate is defensible, NDS returns an explicit unsupported result
rather than silently substituting the nearest food.

\subsection{Access layer}
\label{sec:access}
NDS provides a REST API for applications, named MCP operations for agents, and
Parquet exports for analyses larger than an interactive request. MCP returns
structured records rather than asking agents to extract values from prose.

Responses identify the source system, dataset, and release; food details include
external identifiers, nutrient amounts with units and basis, and portions.
Serving operations are read-only, while ingestion and indexing run
offline.

\subsection{Crosswalk}
\label{sec:crosswalk}

A crosswalk is a versioned set of directed edges between records that lack
shared identifiers. It relates records without merging them or asserting
universal equivalence: both endpoints remain addressable, and each edge states
the relation supported for a declared use and pair of endpoint releases. This
matters because records may describe related foods while differing in
preparation, formulation, geography, analytical basis, or specificity.

\paragraph{Construction.}
Crosswalk construction reuses the matching pipeline, treating a source record
as the query and a target release as the corpus. Parent and child labels provide
context when the target has a hierarchy, and invalid target entities are
removed before reranking. The decision contract returns the target identifier,
relation, confidence, justification, and policy version, or an explicit
no-match decision. Candidate retrieval and relation adjudication remain
separate so that a policy can reject the nearest candidate.

\paragraph{Typed relations.}
Relations are directed from source to target. \emph{exact} denotes the same
food; \emph{broad} means the target is more general; \emph{narrow} means it is
more specific; and \emph{close} denotes related records that do not subsume one
another. \emph{no-match} records that no defensible edge exists. These
distinctions prevent a category or related preparation from being treated
silently as identical and let clients choose relations appropriate to an
analysis.

\paragraph{Release-aware mappings.}
A mapping release is immutable and records its source and target releases,
construction policy, and content identity. Each edge carries stable endpoint
identifiers, relation, confidence, and justification, but does not copy values
from either endpoint. Measurements and mapping decisions therefore remain
separately auditable.

A release becomes visible only through an atomic commit. A workflow pins a
\emph{watermark}, an opaque committed snapshot, and resolution returns the
concrete mapping release and policy selected at that snapshot. Later releases
cannot change a pinned analysis.

\paragraph{Interchange format.}
The runtime representation remains deliberately small; \sssom{} is an exchange
profile rather than the internal query model. Positive edges can use \skos{}
predicates such as \exactmatch{} and \broadmatch{} while retaining endpoint
versions, confidence, mapping justification, provider, and mapping-set
identity~\citep{sssom2022,skos2009}.

\section{Evaluation}
\label{sec:experiments}

We evaluate three operations in an agent-mediated workflow: resolving
descriptions to source records, aligning independent resources, and reusing
those alignments through a pinned agent interface. They test Findability,
Interoperability, and Accessibility/Reusability, respectively; source
reconciliation provides supporting validation.

\subsection{Evaluation: description-driven food matching}
\label{sec:exp-findable}

The first experiment asks whether NDS can turn a free-text meal into
release-specific food evidence. \nutribench~\citep{nutribench2025} evaluates
end-to-end carbohydrate estimation but publishes nutrient targets rather than
the database records underlying each meal. We therefore pair it with an
identifier-preserving set generated from NHANES recalls.

\paragraph{Record-level resolution.}
We generate 1{,}000 held-out meal descriptions from NHANES recalls while
retaining 3{,}597 originating FNDDS codes, similar to the data generation
in~\citep{nutribench2025}. NDS may return a record or abstain;
we pool decisions to calculate precision, recall, and \fone.

\begin{table}[H]
  \centering
  \small
  \caption{Record-level resolution on 1{,}000 held-out meals
  (3{,}597 reference foods). Precision and recall score the top-ranked
  record; recall@5 asks whether the intended record is anywhere in the set
  the reranker accepted.}
  \label{tab:record-resolution}
  \begin{tabular}{@{}lcccc@{}}
    \toprule
    Scoring rule & Recall@5 & Precision & Recall & \fone \\
    \midrule
    Strict source identifier & 0.942 & 0.879 & 0.872 & \textbf{0.875} \\
    Equivalence-aware        & ---   & 0.917 & 0.911 & \textbf{0.914} \\
    \bottomrule
  \end{tabular}
\end{table}

Strict identity \fone is 0.875. The intended record appears among the accepted
candidates for 94.2\% of reference foods but is top-ranked for 87.2\%; much of
the gap is a choice among dense FNDDS variants distinguished by venue,
packaging, or another attribute omitted from the meal text. A secondary
equivalence score credits siblings with matching energy and macronutrients and
raises \fone to 0.914; it does not establish identity or micronutrient
equivalence. The remaining error is therefore driven more by ranking among
near-duplicate records than by failure to retrieve a plausible record.

\paragraph{End-to-end estimation on \nutribench.}
We evaluate all 11{,}857 \nutribench v1 queries. The pipeline follows the
abstention policy in \cref{sec:matching}.

\begin{table}[H]
  \centering
  \small
  \caption{Carbohydrate estimation on all 11{,}857 \nutribench v1 queries.
  Answer rate is the share of queries receiving a supported estimate.
  Acc@7.5\,g is the share of answered queries whose absolute carbohydrate error
  is at most 7.5\,g; mean absolute error (MAE) is also calculated over those
  answers.}
  \label{tab:nutribench}
  \begin{tabular}{@{}lrrrr@{}}
    \toprule
    Condition & Queries & Acc@7.5\,g & MAE (g) & Answer rate \\
    \midrule
    \nds                         & 11{,}857 & \textbf{84.6\%} & \textbf{4.3} & 96.4\% \\
    GPT-4o CoT (best published) & 11{,}857 & 66.8\% & 8.6 & 99.2\% \\
    \bottomrule
  \end{tabular}
\end{table}

\nds answers 96.4\% of queries; among those answers, 84.6\% are within 7.5\,g
and MAE is 4.3\,g, compared with the best published GPT-4o result of 66.8\%
and 8.6\,g. The comparison is unpaired because published per-query outputs are
unavailable. The gain indicates that grounding meal descriptions in source
records improves nutrient estimation, while the unsupported 3.6\%---mostly
foreign meals---exposes a remaining source-coverage limitation.

\subsection{Evaluation: crosswalk}
\label{sec:exp-interop}

The second experiment tests alignment without treating every similar food as
equivalent. We first use an external open-set benchmark, then audit the deployed
FNDDS-to-GI crosswalk.

\paragraph{External crosswalk benchmark: NHANES-to-DFG2}
\citet{lemay2026mapping} publish 1{,}304 NHANES ingredient descriptions
labeled match or no-match against 256 Davis Food Glycopedia 2.0 (DFG2)
reference foods (693 matchable and 611 no-match). We evaluate the contract on
top-25 NDS candidate slates. Before
scoring, we fix the binary projection: \emph{exact}, \emph{broad}, \emph{narrow},
and \emph{close} are matches; abstention is no-match.

\begin{table}[H]
  \centering
  \small
  \caption{Accuracy by label in the NHANES-to-DFG2 benchmark of
  \citet{lemay2026mapping}. Parentheses state the expected decision; $n$ is
  the number of foods.}
  \label{tab:dfg2}
  \begin{tabular}{@{}lrrr@{}}
    \toprule
    Benchmark subset & $n$ & \nds contract & Published system \\
    \midrule
    All labeled foods           & 1{,}304 & \textbf{0.688} & 0.654 \\
    Matchable (link expected)   & 693      & 0.745 & \textbf{0.822} \\
    No-match (abstention expected) & 611   & \textbf{0.624} & 0.466 \\
    \bottomrule
  \end{tabular}
\end{table}

\Cref{tab:dfg2} shows a deliberate precision--recall trade-off. NDS gains 15.8
accuracy points on the 611 no-match foods and loses 7.7 on the 693 matchable
foods, raising overall accuracy from 0.654 to 0.688. Of its 177 errors on
matchable foods, 139 are abstentions; emitted targets agree with the benchmark
93.1\% of the time. Most lost match accuracy therefore comes from refusing a
link, not selecting the wrong target.

Refusals concentrate where the benchmark's matching policy is looser. For
example, it accepts raw broccoli linked to steamed broccoli florets, whereas NDS
avoids the raw-to-cooked transfer and offers frozen broccoli florets. This
protects preparation-sensitive downstream analyses but sacrifices recall under
the benchmark labels. The comparison uses the published aggregate because
per-food outputs are unavailable.

\paragraph{FNDDS-to-GI crosswalk evaluation}
FNDDS reports nutrients and ingredients but no GI. NDS therefore links it to
the 2021 International Tables of Glycemic Index~\citep{atkinson2021gi} through
18{,}222 many-to-many typed edges. We sample 125 deployed edges from each
relation (500 total) and ask an independent LLM judge (Claude Fable 5)
to classify the pair from
the food descriptions and rubric, without seeing the crosswalk decision or GI
value.

\begin{table}[H]
  \centering
  \small
  \caption{Blinded audit of 500 served FNDDS-to-GI mappings, 125 per asserted
  relation. \emph{Defensible} means the judge assigns some relation rather
  than no-match; \emph{as asserted} means the judge's relation equals the
  crosswalk's. Rows are unweighted: every sampled mapping counts equally.}
  \label{tab:gi-audit}
  \begin{tabular}{@{}lccl@{}}
    \toprule
    Asserted relation & Defensible & As asserted & Dominant confusion \\
    \midrule
    \emph{exact}  & 96.8\% & 73.6\% & $\to$ \emph{broad} (16/125) \\
    \emph{broad}  & 92.8\% & 76.8\% & $\to$ \emph{close} (18/125) \\
    \emph{narrow} & 98.4\% & 63.2\% & $\to$ \emph{close} (37/125) \\
    \emph{close}  & 96.8\% & 94.4\% & --- \\
    \midrule
    Unweighted mean & 96.2\% & 77.0\% & --- \\
    \bottomrule
  \end{tabular}
\end{table}

Of the audited mappings, 96.2\% are defensible and 77.0\% receive the asserted
relation. Most disagreement concerns typing rather than whether a mapping
exists: 37 of 125 \emph{narrow} edges are judged \emph{close}, often where one
side specifies a brand and the other a formulation. Judge confidence on these
calls is 0.59, versus 0.78 when it agrees with \emph{narrow}, indicating that
the boundary itself is uncertain. In one case, ``Cake, pound, commercially
prepared, other than all butter, enriched'' maps to ``Pound cake (Sara Lee)'':
the crosswalk calls it \emph{narrow}, while the judge calls it \emph{close}
because one side fixes a formulation and the other a brand. Thus the main
audited weakness is relation granularity, not spurious linking.

Because no complete gold-standard crosswalk exists, we cannot calculate recall
directly. We instead ask a narrower question: when NDS abstains, is it missing a
defensible mapping? The blinded audit contains 200 cases: 125 sampled from the
2{,}058 GI-applicable FNDDS foods without a measured mapping, plus 75 known
mappings whose links are hidden as positive controls. The judge recovers 69 of
75 controls, including all 30 \emph{exact} controls, showing that the audit can
detect clear mappings but is not perfectly sensitive.

Among the 125 actual abstentions, 60 (48\%) have no defensible target and 55
(44\%) have only a dominant-component proxy that the whole-food contract
intentionally rejects. For example, white bread is a relevant component of an
almond-butter sandwich on white bread, but not a mapping for the whole dish. The
remaining 10 (8\%) have a missed \emph{close} or \emph{narrow} target; none has
a missed \emph{exact} target. Thus 115 of 125 abstentions are consistent with
the conservative contract, while the observed misses concern looser relations
rather than exact mappings. These counts are LLM-judged agreement diagnostics,
not a calibrated error rate or an exhaustive recall estimate.

\subsection{Reproducible agent-mediated research}
\label{sec:exp-versioned-crosswalk}

Agents may retrieve different pages or make different joins across runs even
under the same instructions. We test whether moving those choices into a
pinned interface makes the resulting analysis repeatable.

We compute daily glycemic load (GL,
$\sum_i \mathrm{carb}_i\mathrm{GI}_i/100$) for 50 adults sampled from NHANES
2017--2020 day~1 (830 records, 422 distinct foods). The arms receive the same
task and decision rules but different evidence access: a \emph{DIY} arm must
retrieve the 2021 International Tables of Glycemic Index from the open web and
construct the FNDDS-to-GI join, whereas an \emph{NDS} arm resolves the same
foods through MCP against a pinned crosswalk watermark. The frozen DIY prompt
also permits predecessor editions, introducing source-edition latitude. Four
Claude models, ordered from weaker to stronger (Haiku~4.5, Sonnet~5, Opus~5,
and Fable~5), run each arm three times, for 24 isolated runs.
\Cref{tab:gi-person} reports the mean
per-person coefficient of variation (CV) in GL and the share of person--food
pairs assigned the same numeric GI across a model's three repetitions.

\begin{table}[H]
  \centering
  \small
  \caption{Reproducibility by model over three repetitions on the same
  50-person cohort. Lower CV and higher assignment agreement indicate greater
  reproducibility.}
  \label{tab:gi-person}
  \begin{tabular}{@{}llrrrr@{}}
    \toprule
    Metric & Arm & Haiku 4.5 & Sonnet 5 & Opus 5 & Fable 5 \\
    \midrule
    Per-person GL CV (mean) & DIY web & 0.180 & 0.197 & 0.179 & 0.148 \\
                            & NDS MCP & \textbf{0.000} & \textbf{0.000} & \textbf{0.000} & \textbf{0.000} \\
    \addlinespace
    Identical person--food GI & DIY web & 21\% & 64\% & 72\% & 77\% \\
    assignments              & NDS MCP & \textbf{100\%} & \textbf{100\%} & \textbf{100\%} & \textbf{100\%} \\
    \bottomrule
  \end{tabular}
\end{table}

Holding the model fixed does not make the web join reproducible: mean GL CV is
0.148--0.197, and only 21--77\% of person--food assignments remain identical
across a model's three repetitions. In contrast, all 12 NDS
runs return the same 207 numeric food-to-GI assignments, giving every person
the same GL across models and repetitions.

\paragraph{Root causes.} To find where the DIY divergence enters, we audited
the transcripts of Fable~5's three DIY runs. Retrieval nondeterminism is real
--- one run used the prompt's predecessor allowance after failed downloads and
fell back to the 2002 tables
--- but it is not the main cause: the two runs that parsed
\emph{byte-identical} source PDFs into local GI indexes still disagreed on 86
of 422 foods.
\Cref{tab:gi-root-causes} evaluates four potential causes of their divergence.

\begin{table}[H]
  \centering
  \small
  \caption{Why two runs of the same model, holding the same parsed GI tables,
  disagree on 86 of 422 foods. Shares are of the carbohydrate-weighted GL
  divergence between the two runs.}
  \label{tab:gi-root-causes}
  \begin{tabular}{@{}p{0.27\linewidth}rp{0.56\linewidth}@{}}
    \toprule
    Root cause & Share & Example \\
    \midrule
    Same record, opposite applicability ruling & 65\% & Both runs retrieve
    ``Tamales (maize), GI~73'' for the food \emph{tamale with chicken}; one
    assigns 73, the other decides a filled dish is not covered by a
    plain-tamale record and abstains. \\
    \addlinespace
    Different lookup scopes & 18\% & Each run writes its own join procedure: one
    hand-types 91 keyword lookups, the other sweeps regex categories. The
    first never queries \emph{iced tea}, so a GI-72 record sitting in its own
    index goes unused and it abstains on three teas. \\
    \addlinespace
    Different candidate selection & 17\% & ``White rice'' matches 61 published
    records; one run picks 62, the other 72. Spread over 53 foods with a
    median gap of 9 GI points. \\
    \addlinespace
    Record lost in download or parsing & 0\% & No observed disagreement
    was caused by a record being absent from one run's index. \\
    \bottomrule
  \end{tabular}
\end{table}

The largest source of divergence occurs after retrieval: 65\% comes from
opposite judgments about whether the same published record applies to a cohort
food. This is a mapping-policy decision that prompt refinement cannot eliminate
without enumerating each disputed mapping. A further 18\% comes from different
lookup coverage and 17\% from selecting different candidate records. No
observed disagreement is attributable to record loss during download or
parsing. Thus pinning the source document alone is insufficient: the join
procedure and applicability policy must also be fixed. The versioned NDS
crosswalk fixes these decisions before the agent runs, explaining why its
result is invariant. This experiment establishes repeatability for this cohort
and workflow; it does not establish the clinical validity of the selected GI
values.

\subsection{Infrastructure validation}
\label{sec:exp-reusable}

The three experiments above assume that imported values retain their source
meaning. We validate a defined subset of the data stored in NDS, not the full
repository: the listed Foundation, FNDDS, and SR Legacy records and a
deterministic 1{,}000-record sample of Branded foods. Within this subset, we
reconcile nutrient amounts, units, bases, portions, and absent values versus
reported zeros against the source releases.
\Cref{tab:fidelity} reports zero failures across 3.5 million checks in this
evaluated subset.

\begin{table}[H]
  \centering
  \small
  \caption{Record-level source reconciliation for the evaluated subset of four
  USDA FoodData Central releases.}
  \label{tab:fidelity}
  \begin{tabular}{@{}lrrr@{}}
    \toprule
    Release & Foods & Source checks & Failures \\
    \midrule
    Foundation                & 365   & 60{,}301    & 0 \\
    FNDDS 2021--2023          & 5{,}432 & 1{,}289{,}129 & 0 \\
    SR Legacy                 & 7{,}793 & 2{,}096{,}999 & 0 \\
    Branded (1{,}000-record deterministic sample) & 1{,}000 & 54{,}625 & 0 \\
    \midrule
    Total                     & 14{,}590 & 3{,}501{,}054 & 0 \\
    \bottomrule
  \end{tabular}
\end{table}

\section{Limitations}
\label{sec:limitations}

\paragraph{Geographic and source coverage.}
\nds has not yet imported the full range of international and country-level
nutrition datasets. Its coverage outside the United States therefore remains
incomplete.

\paragraph{Licensed and protected data.}
\nds emphasizes public datasets and does not yet mediate subscriptions,
user-specific entitlements, or payment. It has not been evaluated for protected
clinical data, whose consent, privacy, security, and governance requirements
extend beyond this work. Accessibility therefore applies only within source
rights and restrictions; FAIR access does not imply open access.

\paragraph{Evaluation scope and reference quality.}
The evaluation covers selected food-matching and crosswalk tasks, not the full
range of foods or research workflows. Some comparisons rely on published
aggregate results because per-item outputs are unavailable, and several
reference labels are adjudicated by an LLM rather than by domain experts.
Broader evaluation with expert adjudication is therefore still needed.

\paragraph{Evidence quality and downstream responsibility.}
\nds makes sources, versions, and mapping decisions explicit, but it does not
guarantee that the underlying evidence is clinically valid or appropriate for
every analysis. Crosswalks may be incomplete and are not yet systematically
reviewed by nutrition experts. Downstream researchers and agents must therefore
assess fitness for purpose and must not treat NDS outputs as causal, diagnostic,
or therapeutic conclusions.

\section{Conclusion}
\label{sec:conclusion}

Agent-mediated nutrition research requires a new data infrastructure for data
identity, search, and crosswalk. Language models can interpret natural-language
requests, but reproducibility is at risk when every run must rediscover records and
improvise joins. Identity should bind evidence to stable, versioned sources;
search should resolve descriptions to records rather than plausible names; and
crosswalks should state relation type, endpoint versions, provenance, and
abstention conditions.

In the glycemic-load case study, the pinned crosswalk yielded invariant
assignments and glycemic loads across four models and repeated runs, whereas
two web runs using byte-identical source PDFs still disagreed on 86 of 422
foods.

NDS makes this architecture concrete. It provides unified access to
heterogeneous nutrition resources while preserving their source and release
boundaries. It turns food descriptions into traceable record selections and
serves cross-dataset links as typed, versioned mappings rather than opaque joins.
Exposing these operations through an agent-callable interface moves identity,
retrieval, and mapping policy out of individual prompts and into infrastructure
that can be inspected, audited, and replayed.

FAIR principles provide the foundation, but agent use raises the operational
standard. Nutrition infrastructure must do more than return plausible values:
it must make identity, search scope, mapping semantics, uncertainty, and
unsupported operations machine-actionable. This does not replace scientific
judgment. It gives agents a stable, inspectable evidence layer so that researchers can
inspect and reuse the same data choices instead of reconstructing them for every
analysis.

% ---- Bibliography ----
\bibliographystyle{plainnat}
\bibliography{references}

\end{document}